\documentclass[conference]{IEEEtran}
\IEEEoverridecommandlockouts

\usepackage{cite}
\usepackage{amsmath,amssymb,amsfonts}
\usepackage{algorithmic}
\usepackage{graphicx}
\usepackage{textcomp}
\usepackage{xcolor}
\usepackage{url}
\usepackage{pifont}
\usepackage{booktabs}
\usepackage{multirow}
\usepackage{makecell}
\usepackage{array}
\usepackage{ulem}

\def\BibTeX{{\rm B\kern-.05em{\sc i\kern-.025em b}\kern-.08em
    T\kern-.1667em\lower.7ex\hbox{E}\kern-.125emX}}
\begin{document}

\title{FairAdapter: Detecting AI-generated Images with Improved Fairness}

\author{
    \IEEEauthorblockN{1\textsuperscript{st} Feng Ding}\\
    \IEEEauthorblockA{\textit{NanChang University} 
    }
    \and
    \IEEEauthorblockN{2\textsuperscript{nd} Jun Zhang}\\
    \IEEEauthorblockA{\textit{NanChang University}} \\
    \and
    \IEEEauthorblockN{3\textsuperscript{rd} Xinan He$^{\star}$\thanks{$^{\star}$Corresponding author}}\\
    \IEEEauthorblockA{\textit{NanChang University}} \\
    \and
    \IEEEauthorblockN{4\textsuperscript{th} Jianfeng Xu $^{\star}$}\\
    \IEEEauthorblockA{\textit{NanChang University}} \\

    \thanks{This work was supported in part by the National Natural Science Foundation of China under Grant 62262041 and 62266032, and in part by the Jiangxi Provincial Natural Science Foundation under Grant 20232BAB202011, and in part by the Technical Leaders in Major Disciplines-Leading Talents Project under Grant 20225BCI22016.}
}

\maketitle

\begin{abstract}
The high-quality, realistic images generated by generative models
pose significant challenges for exposing them. So far, data-driven deep neural networks have been justified as the most efficient forensics tools for the challenges. However, they may be over-fitted to certain semantics, resulting in considerable inconsistency in detection performance across different contents of generated samples. It could be regarded as an issue of detection fairness. In this paper, we propose a novel framework named Fairadapter to tackle the issue. In comparison with existing state-of-the-art methods, our model achieves improved fairness performance. Our project: \url{https://github.com/AppleDogDog/FairnessDetection}
\end{abstract}


\section{Introduction}
With the constant development of artificial intelligence, numerous generative models can synthesize high-quality images that are visually indistinguishable \cite{b1,b2}. In particular, large generative models are capable of generating corresponding images based on any textual description, enabling individuals without specialized knowledge to massively generate high-quality images. People with ulterior motives may use these technologies to create images containing false information to spread on the internet, which poses a great threat to cyberspace.

To address these challenges, researchers have explored various data-driven models for forensic purposes \cite{b3, b4, b5, b6}, including large visual-language model CLIP (Contrastive Language–Image Pre-training)\cite{b7}.

However, the performance of such detectors relies heavily on large volumes of training data. During data collection, the representation of different groups or categories may be incomplete or imbalanced. Images of certain groups may be significantly underrepresented compared to others in the training set. If the data harbor biases, the model may inherit and amplify them. Therefore, pre-trained forensics tools fail to consistently maintain stable performance across all image contents, which yields poor fairness. 

To tackle this, we introduce a novel framework, termed FairAdapter, along with a corresponding training strategy designed to develop fair performance for detecting AI-generated non-facial images.

Our contributions can be summarized as follows: 

\begin{itemize}
    \item  We pioneer in investigating the fairness issue in the context of AI-generated non-facial image detection.
    \item we propose the FairAdapter, combined with a custom-designed learning strategy and categories loss function to tackle the fairness issue raised in detecting AI-generated images.
    \item  Experimental results show that our proposed model could achieve superior fairness performance in predicting AI-generated images. Also, the merit of the proposed modules is identified via ablation studies.
\end{itemize}

\section{RELATED WORKS}
\subsection{Image generation models}
Deep neural networks have achieved significant maturity in the domain of image generation. 
At first, GANs exhibiting exceptional performance in synthesizing images are employed for generative tasks \cite{b8,b9}. 
In recent years, the advent of diffusion models has brought text-to-image generation methods to the forefront of public attention\cite{b10}. They generate high-quality images by iteratively adding and then reversing noise to data, learning to reconstruct an image from a noisy version through a denoising process, and effectively capturing complex data distributions.

\subsection{Forensics for AI-generated images}
With the emergence of numerous advanced generative models, research on detecting images produced by these models has become increasingly significant.
Zhang \emph{et al.}\cite{b11} disclosed the artifacts induced during the up-sampling of the auto-encoder in the frequency domain. They designed a forensics method to trace the artifacts for exposing GAN-generated images.
Wang \emph{et al.}\cite{b3} proposed to train a deep neural network that can effectively detect the image generated by various GANs. 
In recent years, Ojha \emph{et al.}\cite{b7} employed a frozen CLIP model to train a linear classifier to detect images generated by GANs as well as diffusion models. The CLIP model is also adopted by Khan \emph{et al.}\cite{b12} for achieving high forensics performance in discerning AI-generated images.
Despite the great progress made so far, 
existing forensics technologies are often prone to issues of data bias and unfairness\cite{b13}.
In deepfake detection, some research has addressed fairness issues related to demographic semantics\cite{b14,b13}. However, research about fairness in the domain of AI-generated non-facial image detection remains limited.

\section{METHODOLOGY}
\subsection{Motivation}
According to published reports~\cite{b3,b7,b12}, the forensics models for detecting AI-generated images may be overfitted to certain semantics. Such overfitting may facilitate increased inconsistency across different contents. This phenomenon happens because the detection capability of data-driven detectors relies heavily on the semantics embedded in the training data. With biases in training datasets, they may struggle with fairly detecting cross-domain contents. 

Therefore, the proposed method is designated to constantly search for the traces left by generative models while easing the impact of irrelevant semantics. This strategy enables the detector to accurately identify the traces of generative models within a more intricate semantic context, thereby mitigating the model's bias toward specific semantics.

\begin{figure*}[t]
    \centering    
    \includegraphics[width=1.0\textwidth]{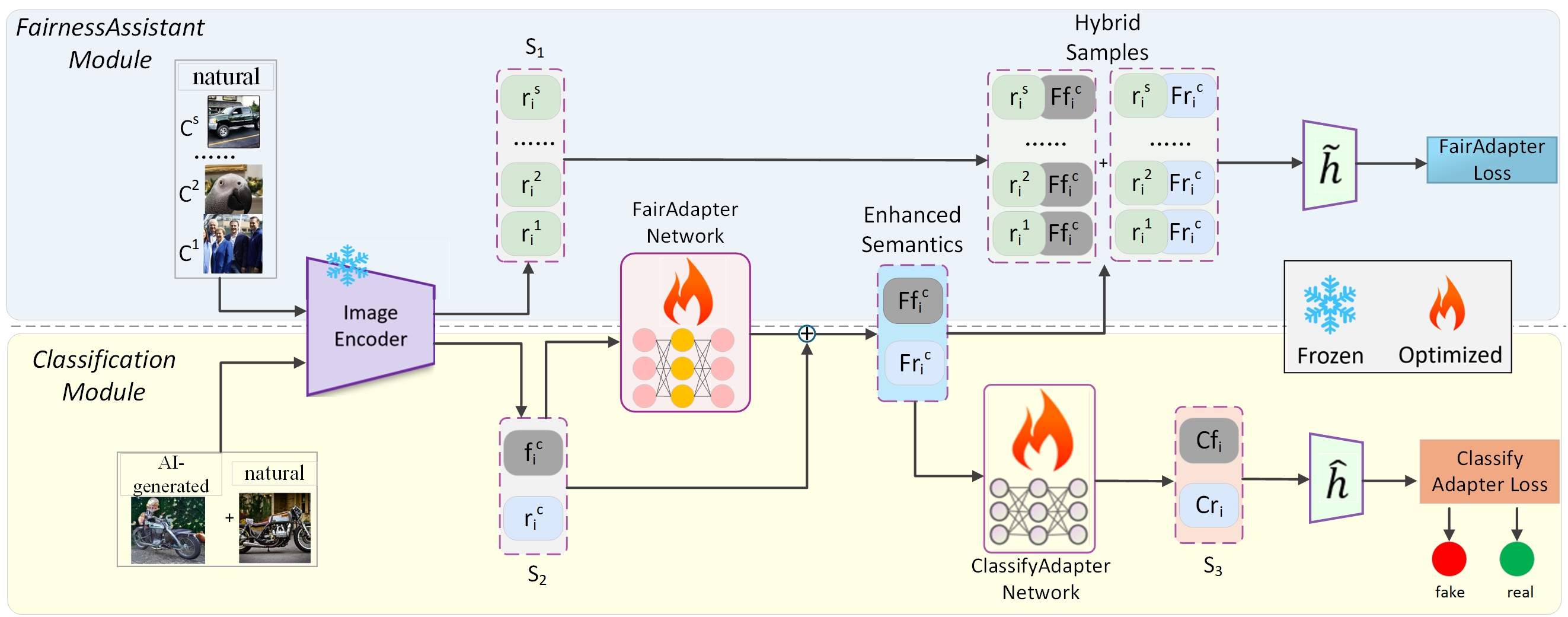} 
    \caption{An overview of our proposed method: 1) For the input stage, we select one category of natural and AI-generated images, along with natural images of other categories, to input into the image encoder. 2) In the FairnessAssistant module, we compute the FairAdapter loss by mixing the image semantics. 3) In the Classification module, we feed the enhanced semantics into the ClassifyAdapter network to complete the classification task.}
    \label{fig:overview}
\end{figure*} 

\subsection{Overview of Proposed Method}
Fig.~\ref{fig:overview} shows the framework of our approach, comprising three modules: fairness assistant module, classification module, and loss function. The fairness assistant module is a pre-trained CLIP image encoder proposed in \cite{b15}. It is designed to extract natural semantics ($i.e.$,$s_1$) from images of other categories to assist the classification module. The classification module, primarily used to extract semantics ($i.e.$,$s_3$) for classification, is composed of the same image encoder, the FairAdapter network, and the ClassifyAdapter network.
Both adapter networks share the same structure consisting of two fully connected layers and are trained independently.

We use an image encoder ($i.e.$, $\mathbf{E}(\cdot)$) to extract the original image semantics($i.e.$,$s_2$) that are the key features used to determine whether an image is AI-generated from image pairs $(X_i^c, {X^c}^{'}_i)$. $X_i^c$ denotes AI-generated image and ${X^c}^{'}_i$ denotes natural image, with $c \in \{1,\cdots,n\}$ represent categories. Meanwhile, we also use the same Image Encoder to extract semantics from other images that are different categories from the AI-generated images to obtain $r_i^s$, and $c\notin s$:

\begin{align}
    &f_i^c, r_i^c=\mathbf{E}(X_i^c, {X^c}^{'}_i),\! \\
    &r_i^s = \mathbf{E}(C^s),
\end{align}
where $r_i^c$ and $f_i^c$ represent the original image semantics($s_2$) of natural semantics and AI-generated semantics by processed Image Encoder.

\subsubsection{FairAdapter Network}
We pass the original image semantics($s_2$) through our FairAdapter network $\mathbf{A_f}$ to obtain enhanced semantics and combine them with natural semantics($s_1$) to generate hybrid samples ($i.e.$, $f^{mix}_i$), which consists of a single category of enhanced semantics and original image semantics($s_2$) from other categories. Then, we will use the hybrid samples within our designed FairAdapter loss(ref to Sec~\ref{sec:Categories Loss Function}).

\begin{align}
    &Ff^c_i, Fr_i^c= \mathbf{A_f}(f_i^c, r_i^c) + (f_i^c, r_i^c)  \\
    &f_i^{mix} = cat((Ff^c_i / Fr_i^c), r_i^s),
\end{align}
where $Ff^c_i$ and $Fr_i^c$ represent the AI-generated enhanced semantics and natural enhanced semantics, respectively. $cat$ denotes tensor concatenation.

\subsubsection{ClassifyAdapter network}
We use ClassifyAdapter network $\mathbf{A_c}$ to extract the enhanced final image classification semantics($s_3$) of enhanced semantics.

\begin{align}
    Cf_i, Cr_i= \mathbf{A_c}(Ff^c_i, Fr_i^c),
\end{align}
where $Cf_i$ and $Cr_i$ represent the enhanced final image classification semantics. 

\subsubsection{Categories Loss Function}
\label{sec:Categories Loss Function}
The categories loss function comprises two components, the FairAdapter Loss for the optimized FairAdapter network and the ClassifyAdapter Loss for the optimized ClassifyAdapter network.

\paragraph{FairAdapter Loss}
We use the FairAdapter loss to penalize the predictive values of hybrid samples from both AI-generated and natural images across each category. During loss computation, a separate cross-entropy loss $\mathbf{CE}(\cdot,\cdot)$ is calculated for each category, followed by dynamic adjustments to improve performance. Specifically, we record the previous loss values and dynamically compute a weight for each category based on both current and past loss values.The detailed process is as follows,

\begin{align}
    &L_1,\cdots,L_j = \mathbf{CE}(\widetilde{h}(f_i^{mix}),Y_i)  \\
    &L_{fair} = \frac{\sum_{i=1}^{n}\lambda_jL_j}{n},
\end{align}

where $\widetilde{h}$ are the classification heads for $f_i^{mix}$, $Y_i$ is the classification label denoting fake and real. $L_j$ represents the loss of images in the corresponding category ($j\in \{1,\cdots,n\}$), and $L_{fair}$ represents FairAdapter loss. $\lambda_j$ is dynamically calculates the weight for updating the corresponding category loss. The calculation process is as follows:

\begin{align*}
    \left\{\begin{matrix} 
        \lambda_j =1 + \frac{L_j}{L_j^{pre}}&\quad  L_j^{pre} < L_j\\  
        \lambda_j =1 - \frac{L_j^{pre}}{L_j}&\quad  L_j^{pre} \ge L_j,
    \end{matrix}\right. 
\end{align*}
where $L_j^{pre}$ corresponds to the last calculated loss value of this category. 

\paragraph{ClassifyAdapter Loss}
We use ClassifyAdapter loss to learn the classification features of a pair of positive and negative samples processed by the ClassifyAdapter network. It can be presented as follows:

\begin{align}
    &L_c = \mathbf{CE}(\widehat{h}(Cf_i),Y_i),
\end{align}
where $\widehat{h}$ denotes the classification header of $Cf_i$,

\section{EXPERIMENTS}
\subsection{Experimental Settings}

\textbf{Dataset.}
To examine the performance of the proposed model, all models are trained with the ProGAN dataset~\cite{b3}. The training set consists of 20 categories of different image contents. Each category contains 300 natural images and 300 AI-generated images. The total number of training images is 12k. 
All trained models are tested on ProGAN\cite{b16} for intra-domain detection evaluation, and tested on CycleGAN\cite{b17}, StyleGAN\cite{b18}, and StyleGAN2\cite{b19} datasets for cross-domain evaluations. In addition, we test the datasets generated by the diffusion model, including ldm\cite{b10}, Glide\cite{b20}, Stable diffusion\cite{b10}, to evaluate the generalizability of the models.

\begin{table}[]
\caption{Detection Performance on ProGAN}
\centering
\scalebox{1}{
    \resizebox{\columnwidth}{!}{%
        \begin{tabular}{clll|ccc}
            \hline
            \multicolumn{4}{c|}{}     & \multicolumn{3}{c}{progan(\%)}                                                                            \\ \cline{5-7} 
            \multicolumn{4}{c|}{method}                            & \multicolumn{2}{c|}{Fairness Metrics$\downarrow$}           & \begin{tabular}[c]{@{}c@{}}Detection\\ Metric$\uparrow$\end{tabular} \\ \cline{5-7} 
            \multicolumn{4}{c|}{}                            & $F_{AUC}$       & \multicolumn{1}{c|}{$F_{FPR}$}    & AUC                                                     \\ \hline
            \multicolumn{4}{c|}{CNNspot\cite{b3}}         & 12.97         & \multicolumn{1}{c|}{23.0}         & 71.36                                                      \\ 
            \multicolumn{4}{c|}{FreDect\cite{b21}} & {\underline{0.02}}    & \multicolumn{1}{c|}{0.99} & {99.99}                                                \\ 
            \multicolumn{4}{c|}{Fusing\cite{b22}}         & 0.06          & \multicolumn{1}{c|}{3.50}        & 99.98                                                      \\ 
            \multicolumn{4}{c|}{Dmimg\cite{b4}}  & 10.18         & \multicolumn{1}{c|}{22.50}       & 58.90                                                       \\ 
            \multicolumn{4}{c|}{Lgrad\cite{b23}}     & \textbf{0.01} & \multicolumn{1}{c|}{{\underline{2.0}}}    & {\underline{99.99}}                                                \\ \hline
            \multicolumn{4}{c|}{CLIP+Linear Probing\cite{b7}}        & 0.25          & \multicolumn{1}{c|}{3.0}         & 99.79                                                     \\ 
            \multicolumn{4}{c|}{CLIP+Adapter\cite{b24}}         & 1.13          & \multicolumn{1}{c|}{5.0}          & 99.85                                                      \\ \hline
            \multicolumn{4}{c|}{CLIP+FairAdapter(ours)}                        & 0.53          & \multicolumn{1}{c|}{\textbf{1.0}} & \textbf{100}                                               \\ \hline
        \end{tabular}%
    }
} 
\label{tab:domain}
\end{table}

\begin{table*}[]
\caption{In the cross-domain scenario (CycleGAN,StyleGAN,StyleGAN2), it is compared with different methods in terms of fairness and generalization. ↑ means higher is better and ↓ means lower is better.}
\centering
\scalebox{2}{
    \resizebox{\columnwidth}{!}{%
        \begin{tabular}{cll|ccc|ccc|ccc}
            \hline
            \multicolumn{3}{c|}{} & \multicolumn{3}{c|}{cyclegan(\%)}                                                                               & \multicolumn{3}{c|}{stylegan(\%)}                                                                              & \multicolumn{3}{c}{stylegan2(\%)}                                                                             \\ \cline{4-12} 
            \multicolumn{3}{c|}{method}                        & \multicolumn{2}{c|}{Fairness Metrics↓}              & \begin{tabular}[c]{@{}c@{}}Detection\\ Metric↑\end{tabular} & \multicolumn{2}{c|}{Fairness Metrics↓}             & \begin{tabular}[c]{@{}c@{}}Detection\\ Metric↑\end{tabular} & \multicolumn{2}{c|}{Fairness Metrics↓}             & \begin{tabular}[c]{@{}c@{}}Detection\\ Metric↑\end{tabular} \\ \cline{4-12} 
            \multicolumn{3}{c|}{}                        & $F_{AUC}$       & \multicolumn{1}{c|}{$F_{FPR}$}       & AUC                                                     & $F_{AUC}$       & \multicolumn{1}{c|}{$F_{FPR}$}      & AUC                                                     & $F_{AUC}$       & \multicolumn{1}{c|}{$F_{FPR}$}      & AUC                                                     \\ \hline
            \multicolumn{3}{c|}{CNNspot~\cite{b3}}                 & 17.86         & \multicolumn{1}{c|}{40.53}         & 69.60                                                       & 10.28         & \multicolumn{1}{c|}{10.87}        & 64.80                                                       & 4.48          & \multicolumn{1}{c|}{12.12}        & 68.18                                                      \\ 
            \multicolumn{3}{c|}{FreDect~\cite{b21}}                 & 33.61         & \multicolumn{1}{c|}{11.65}         & 87.76                                                      & 32.46         & \multicolumn{1}{c|}{34.75}        & 87.41                                                      & 26.86         & \multicolumn{1}{c|}{28.84}        & 89.32                                                      \\ 
            \multicolumn{3}{c|}{Fusing~\cite{b22}}                  & 5.22          & \multicolumn{1}{c|}{6.72}          & 96.18                                                      & 9.31          & \multicolumn{1}{c|}{\textbf{0.40}} & 92.66                                                      & 11.72         & \multicolumn{1}{c|}{\textbf{0.50}} & 91.43                                                      \\ 
            \multicolumn{3}{c|}{Dmimg~\cite{b4}}                   & 40.13         & \multicolumn{1}{c|}{99.36}         & 43.78                                                      & 17.99         & \multicolumn{1}{c|}{95.09}        & 63.10                                                       & 27.61         & \multicolumn{1}{c|}{95.09}        & 80.76                                                      \\ 
            \multicolumn{3}{c|}{LGrad~\cite{b23}}                   & 25.23         & \multicolumn{1}{c|}{11.05}         & 91.84                                                      & 13.67         & \multicolumn{1}{c|}{6.96}         & 95.42                                                      & 10.48         & \multicolumn{1}{c|}{3.46}         & \textbf{97.10}                                              \\ \hline
            \multicolumn{3}{c|}{CLIP+Linear Probing~\cite{b7}}                    & {\underline {1.26}}    & \multicolumn{1}{c|}{16.24}         & \textbf{99.32}                                             & 7.79          & \multicolumn{1}{c|}{1.05}         & 95.63                                                      & {\underline {1.58}}    & \multicolumn{1}{c|}{1.25}         & 92.57                                                      \\ 
            \multicolumn{3}{c|}{CLIP+Adapter~\cite{b24}}                 & \textbf{1.16} & \multicolumn{1}{c|}{{\underline {9.77}}}    & {\underline {99.01}}                                                & {\underline {4.39}}    & \multicolumn{1}{c|}{1.30}          & \underline {97.66}                                                      & 3.32          & \multicolumn{1}{c|}{1.35}         & 94.17                                                      \\ \hline
            \multicolumn{3}{c|}{CLIP+FairAdapter(ours)}                    & 1.85          & \multicolumn{1}{c|}{\textbf{1.26}} & 98.80                                                       & \textbf{4.04} & \multicolumn{1}{c|}{{\underline {1.05}}}   & \textbf{97.83}                                             & \textbf{1.51} & \multicolumn{1}{c|}{{\underline {0.95}}}   & {\underline {96.87}}                                                \\ \hline
        \end{tabular}%
    }
}
\label{tab:generalization}
\end{table*}

\begin{table*}[]
\caption{We studied the ablation of two key modules. "FairAdapter" represents the FairAdapter Network we proposed. "fCls(Fair)" denotes the Adapter Loss for the optimized FairAdapter network. "cCls(noFair)" refers to the ClassifyAdapter Loss that is not optimized for the FairAdapter network.}
\centering
\scalebox{2}{
    \resizebox{\columnwidth}{!}{%
        \begin{tabular}{cccc|cccccccc}
        \hline
        \multicolumn{4}{c|}{method}                        & \multicolumn{8}{c}{dataset($\%$)}                                                                                                                                                            \\ \cline{5-12} 
        \multicolumn{4}{c|}{}                                               & \multicolumn{2}{c|}{progan}                    & \multicolumn{2}{c|}{cyclegan}                       & \multicolumn{2}{c|}{stylegan}                       & \multicolumn{2}{c}{stylegan2} \\ \hline
        \multicolumn{1}{c|}{name}     & \multicolumn{1}{c}{FairAdapter} & fCls(Fair) & cCls(noFair) &$F_{FPR}$$\downarrow$    & \multicolumn{1}{c|}{AUC$\uparrow$}       & $F_{FPR}$$\downarrow$       & \multicolumn{1}{c|}{AUC$\uparrow$}         & $F_{FPR}$$\downarrow$       & \multicolumn{1}{c|}{AUC$\uparrow$}         & $F_{FPR}$$\downarrow$       & AUC$\uparrow$         \\ \hline
        
        \multicolumn{1}{c|}{VariantA} & \multicolumn{1}{c}{}        &     & \ding{51}& 5.0          & \multicolumn{1}{c|}{{ 99.85}}  & {9.77}    & \multicolumn{1}{c|}{{99.01}}    & {1.30}     & \multicolumn{1}{c|}{{97.66}}    & {1.35}    & {\underline{94.17}}    \\ 
        
        \multicolumn{1}{c|}{VariantB} & \multicolumn{1}{c}{\ding{51}}     &  &    \ding{51}  & {\underline{1.50}}  & \multicolumn{1}{c|}{95.24}        & \underline{1.27}          & \multicolumn{1}{c|}{\textbf{99.15}} & 1.45          & \multicolumn{1}{c|}{97.22}          & 1.50           & 93.48          \\  
       \multicolumn{1}{c|}{VariantC} & \multicolumn{1}{c}{\ding{51}}     & \ding{51}  &     & {2.0}  & \multicolumn{1}{c|}{\underline{99.93}}        &  2.26          & \multicolumn{1}{c|}{\underline{99.02}} & \textbf{0.95}          & \multicolumn{1}{c|}{\textbf{98.07}}          &\underline{1.0}           & 93.58          \\  \hline
        
        \multicolumn{1}{c|}{ours}     & \multicolumn{1}{c}{\ding{51}}       & \ding{51} &   \ding{51} & \textbf{1.0} & \multicolumn{1}{c|}{\textbf{100}} & \textbf{1.26} & \multicolumn{1}{c|}{98.80}           & \underline{1.05} & \multicolumn{1}{c|}{\underline{97.83}} & \textbf{0.95} & \textbf{96.87} \\ \hline
        \end{tabular}%
    }
}
\label{tab:Ablation}      
\end{table*}

\textbf{Evaluation Metrics.}
To evaluate the detection performance, we employ the Area Under the Curve (AUC) which averages the AUCs for predicting images of different contents. we adopt FAUC, FFPR\cite{b13} to evaluate the fairness performance. We assume a test set comprising indices $\{1, \ldots, n\}$. Let $Y_j$ and $\hat{Y}_j$ represent the true and predicted labels of the sample $X_j$, respectively. Both $Y_j$ and $\hat{Y}_j$ are binary values, where $0$ denotes "natural" and $1$ denotes "AI-generated". For all fairness metrics, a lower value indicates better performance. These metrics are defined as follows:

\begin{align*}
    &F_{F P R}:=\sum_{\mathcal{J}_j \in \mathcal{J}}\left|\frac{\sum_{j=1}^n \mathbb{I}_{\left[\hat{Y}_j=1, D_j=\mathcal{J}_j, Y_j=0\right]}}{\sum_{j=1}^n \mathbb{I}_{\left[D_j=\mathcal{J}_j, Y_j=0\right]}}-\frac{\sum_{j=1}^n \mathbb{I}_{\left[\hat{Y}_j=1, Y_j=0\right]}}{\sum_{j=1}^n \mathbb{I}_{\left[Y_j=0\right]}}\right|, \\
    &F_{A U C} := \max_{\mathcal{J}_j \in \mathcal{J}} \left\{
    \frac{\sum_{j=1}^n \mathbb{I}_{\left[\hat{Y}_j=Y_j, D_j=\mathcal{J}_j\right]}}
    {\sum_{j=1}^n \mathbb{I}_{\left[D_j=\mathcal{J}_j\right]}} \right. \\
    &\quad\quad\quad\quad\quad\quad\quad - \left. \min_{\mathcal{J}_j^{\prime} \in \mathcal{J}} 
    \frac{\sum_{j=1}^n \mathbb{I}_{\left[\hat{Y}_j=Y_j, D_j=\mathcal{J}_j^{\prime}\right]}}
    {\sum_{j=1}^n \mathbb{I}_{\left[D_j=\mathcal{J}_j^{\prime}\right]}} 
    \right\},
\end{align*}

where \( D \) is the contents variable, \( \mathcal{J} \) is the set of subgroups with each subgroup \( \mathcal{J}_j \in \mathcal{J} \). \( F_{FPR} \) measures the disparity in False Positive Rate (FPR) across different groups compared to the overall contents.\( F_{AUC} \) measures the maximum AUC gap across all demographic groups.

\textbf{Experimental details.}
Other than conventional convolutional neural networks\cite{b3,b21,b22,b4,b23}, we also employ two methods based on CLIP\cite{b7,b24} to conduct performance comparisons.
All models are built with Pytorch and trained on a server with an NVIDA RTX 4090. For training, we fixed the epoch to 40, the batch size of the baseline method to 32, our method to 1, the ADM optimizer, and the learning rate to 0.0002. Some benchmark models are tested with the weight released by authors. 

\subsection{Experimental Results}
\textbf{Performance of intra-domain.}
At first, we test all the models on the ProGAN test set to conduct intra-domain evaluations. 
As shown in Table \ref{tab:domain}, our method generally exhibits improved fairness over the compared benchmarks and achieves higher AUCs on most methods. The results demonstrate that eliminating interfering semantics and concentrating on fake semantics can enhance the model's detection performance.

\textbf{Performance of cross-domain.}
Table \ref{tab:generalization} shows that our method could outperform existing state-of-the-art benchmarks in most cases. Specifically, compared with traditional CNN detection methods\cite{b3}, our method increases FAUC by 16.01\% on CycleGAN, 9.82\% on StyleGAN, and 11.17\% on StyleGAN2.
Furthermore, as shown in Table \ref{tab:diffusion}, even when applied to the dataset generated by the diffusion model, our method demonstrates superior fairness and detection performance compared to previous approaches.

\begin{table}[]
\caption{Cross-domain evaluation on the dataset generated by the diffusion model. Since the dataset does not contain multiple categories, we use FPR as the fairness indicator and AUC as the detection indicator}
\centering
\resizebox{\columnwidth}{!}{%
    \begin{tabular}{cll|cccccccc}
        \hline
        \multicolumn{3}{c|}{} & \multicolumn{8}{c}{dataset}                                                                                                                                                                   \\ \cline{4-11} 
        \multicolumn{3}{c|}{method}                        & \multicolumn{2}{c|}{ldm}                           & \multicolumn{2}{c|}{glide}                          & \multicolumn{2}{c|}{sd1\_4}                         & \multicolumn{2}{c}{sd1\_5}   \\ \cline{4-11} 
        \multicolumn{3}{c|}{}                        & FPR↓          & \multicolumn{1}{c|}{AUC↑}            & FPR↓           & \multicolumn{1}{c|}{AUC↑}            & FPR↓           & \multicolumn{1}{c|}{AUC↑}            & FPR↓          & AUC↑            \\ \hline
        \multicolumn{3}{c|}{CNNspot~\cite{b3}}                 & 40.50         & \multicolumn{1}{c|}{68.35}          & 47.25         & \multicolumn{1}{c|}{71.81}          & 58.65         & \multicolumn{1}{c|}{40.88}          & 47.51        & 59.91          \\ 
        \multicolumn{3}{c|}{FreDect~\cite{b21}}                 & 34.50         & \multicolumn{1}{c|}{92.36}          & 33.40          & \multicolumn{1}{c|}{60.92}          & 33.17         & \multicolumn{1}{c|}{30.19}          & 32.31        & 30.10           \\ 
        \multicolumn{3}{c|}{Fusing~\cite{b22}}                  & 2.10          & \multicolumn{1}{c|}{70.62}          & \underline {1.27}          & \multicolumn{1}{c|}{63.72}          & 0.97          & \multicolumn{1}{c|}{60.05}          & \textbf{1.04}         & 58.62          \\ 
        \multicolumn{3}{c|}{Dmimg~\cite{b4}}                   & 84.10         & \multicolumn{1}{c|}{42.28}          & 91.06         & \multicolumn{1}{c|}{56.86}          & 90.54         & \multicolumn{1}{c|}{41.54}          & 90.46        & 41.60           \\ 
        \multicolumn{3}{c|}{Lgrad~\cite{b23}}                   & 4.80          & \multicolumn{1}{c|}{96.18}          & 35.25         & \multicolumn{1}{c|}{79.87}          & 36.68         & \multicolumn{1}{c|}{\textbf{68.96}} & 35.65        & \textbf{69.37} \\ \hline
        \multicolumn{3}{c|}{CLIP+Linear Probing~\cite{b7}}                    & 3.40          & \multicolumn{1}{c|}{\textbf{97.69}} & 3.66          & \multicolumn{1}{c|}{76.77}          & 4.37          & \multicolumn{1}{c|}{61.45}          & 3.80          & 61.79          \\ 
        \multicolumn{3}{c|}{CLIP+Adapter~\cite{b24}}                 & {\underline {1.40}}    & \multicolumn{1}{c|}{{\underline {97.62}}}    & 2.34          & \multicolumn{1}{c|}{\underline {81.80}}           & \underline {2.51}          & \multicolumn{1}{c|}{{\underline {68.49}}}    & 2.40          & {\underline {69.0}}       \\ \hline
        \multicolumn{3}{c|}{CLIP+FairAdapter(ours)}                    & \textbf{0.40} & \multicolumn{1}{c|}{97.21}          & \textbf{0.91} & \multicolumn{1}{c|}{\textbf{86.63}} & \textbf{1.06} & \multicolumn{1}{c|}{63.96}          & \underline {1.30} & 64.39          \\ \hline
    \end{tabular}%
}
\label{tab:diffusion}
\end{table}

\textbf{Ablation Study.}
The performance of our proposed method is mainly affected by three components: the FairAdapter network, the FairAdapter Loss, and the ClassifyAdapter Loss. Therefore, we carried out ablation experiments on each component, corresponding to the results of variables A, B, and C in Table \ref{tab:Ablation}, respectively. For example, observed from the experimental results, when the FairAdapter network is removed, the fairness performance is inferior with an increase of 4\% for FFPR, while the detection performance degrades by a decrease of 0.15\% for AUC. 

\section{Conclusion}
\label{sec:majhead}

In this paper, we propose a method for detecting AI-generated images. Unlike traditional approaches, our method not only addresses the detection challenge but also emphasizes the fairness issues inherent in the process. Leveraging the robust semantic analysis capabilities of CLIP, our proposed FairAdapter network effectively mitigates the impact of interfering semantics. Experimental validation demonstrates that our method achieves strong detection performance, and more importantly, it successfully addresses the issue of detection fairness.

One of the limitations of our method is that it can not well detect the images generated across the model, that is, after ProGAN training, the accuracy of the images generated by the diffusion model will be reduced. In the future, we will strive to conduct relevant research to address this issue.


\end{document}